\documentclass[10pt,twocolumn,letterpaper]{article}

\usepackage[pagenumbers]{cvpr}

\usepackage{multirow}
\usepackage[table]{xcolor}
\definecolor{lightgray}{gray}{0.9}
\definecolor{myblue}{HTML}{DFF2FC}
\definecolor{secondbest}{HTML}{FFF8AE}
\definecolor{thirdbest}{HTML}{FFCC99}
\usepackage{placeins}

\definecolor{cvprblue}{rgb}{0.21,0.49,0.74}
\usepackage[pagebackref,breaklinks,colorlinks,allcolors=cvprblue]{hyperref}

\def\paperID{20874} 
\def\confName{CVPR}
\def\confYear{2026}

\title{SPyCer: Semi-Supervised Physics-Guided Contextual Attention for Near-Surface Air Temperature Estimation from Satellite Imagery}

\author{
\makebox[\linewidth]{%
\parbox{\linewidth}{\centering
Sofiane Bouaziz$^{1,2}$, Adel Hafiane$^2$, Raphaël Canals$^2$, Rachid Nedjai$^3$\\[2mm]
$^1$INSA CVL, Université d’Orléans, PRISME UR 4229, Bourges, 18022, Centre Val de Loire, France\\
$^2$Université d’Orléans, INSA CVL, PRISME UR 4229, Orléans, 45067, Centre Val de Loire, France\\
$^3$Université d'Orléans, CEDETE, UR 1210, Orléans, 45067, Centre Val de Loire, France
}}%
}

\begin{document}
\maketitle
\begin{abstract}

Modern Earth observation relies on satellites to capture detailed surface properties. Yet, many phenomena that affect humans and ecosystems unfold in the atmosphere close to the surface. Near-ground sensors provide accurate measurements of certain environmental characteristics, such as near-surface air temperature (NSAT). However, they remain sparse and unevenly distributed, limiting their ability to provide continuous spatial measurements. To bridge this gap, we introduce SPyCer, a semi-supervised physics-guided network that can leverage pixel information and physical modeling to guide the learning process through meaningful physical properties. It is designed for continuous estimation of NSAT by proxy using satellite imagery. SPyCer frames NSAT prediction as a pixel-wise vision problem, where each near-ground sensor is projected onto satellite image coordinates and positioned at the center of a local image patch. The corresponding sensor pixel is supervised using both observed NSAT and physics-based constraints, while surrounding pixels contribute through physics-guided regularization derived from the surface energy balance and advection-diffusion-reaction partial differential equations. To capture the physical influence of neighboring pixels, SPyCer employs a multi-head attention guided by land cover characteristics and modulated with Gaussian distance weighting. Experiments on real-world datasets demonstrate that SPyCer produces spatially coherent and physically consistent NSAT estimates, outperforming existing baselines in terms of accuracy, generalization, and alignment with underlying physical processes.

\end{abstract}    
\section{Introduction}

\begin{figure}[t]
  \centering
  \includegraphics[width=0.47\textwidth]{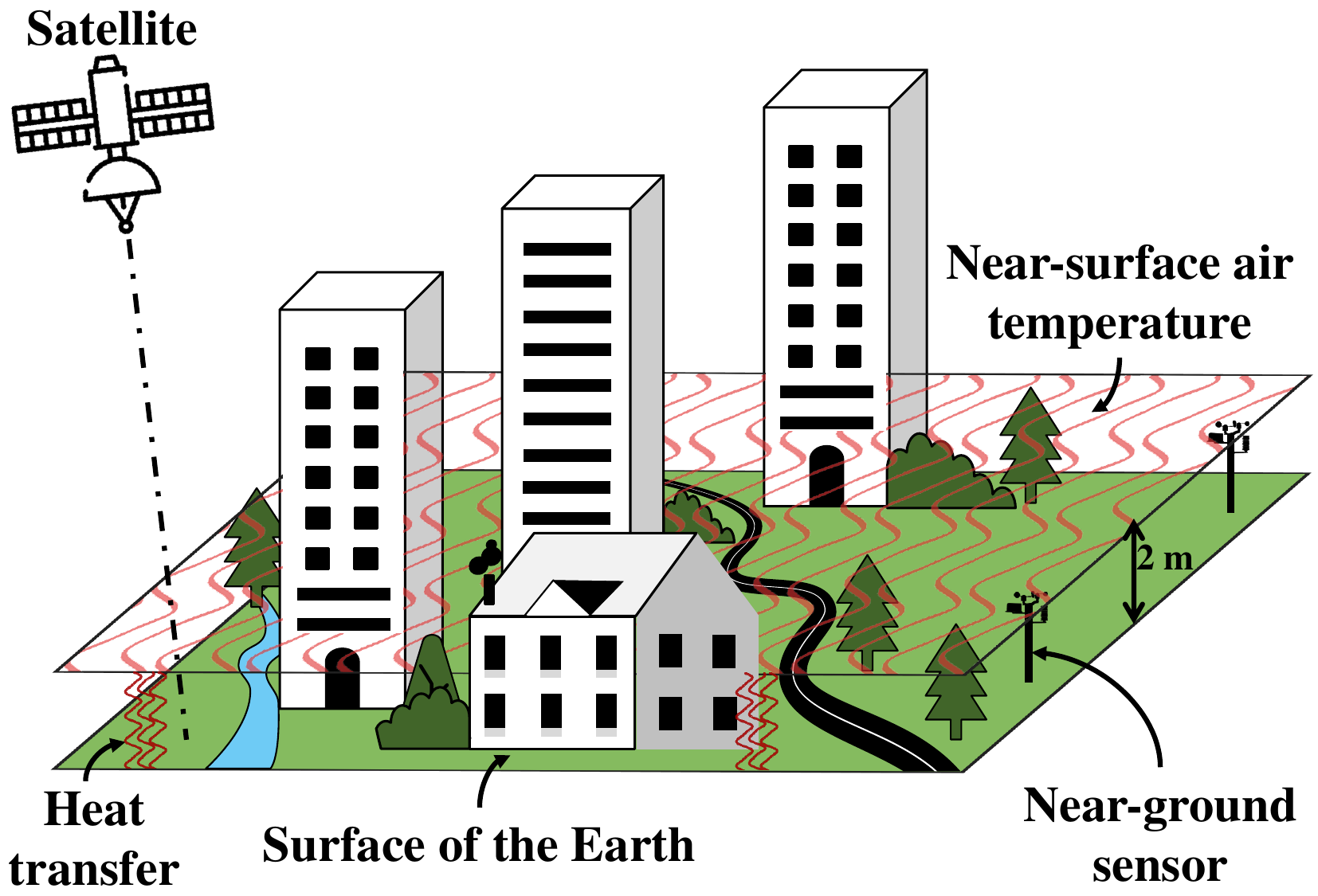}
  \caption{Illustration of the sensing gap between satellite observations and near-ground conditions. Satellites capture surface properties, while NSAT, measured 2 meters above the ground, drives human comfort and environmental processes. Near-ground sensors provide accurate but sparse measurements, leaving most areas unsampled. SPyCer leverages the energy exchange between land surface and near-surface atmosphere to estimate continuous, physically consistent NSAT from sparse sensors and satellite imagery.}
  
  \label{fig:problem_illustration}
\end{figure}

Satellite imagery transformed how we monitor the Earth by capturing key physical variables at meter-scale spatial resolution~\cite{zhang2022artificial, ghamisi2025responsible, lu2025vision}. Among these, land surface temperature (LST) reveals detailed patterns of surface heating~\cite{li2013satellite}. However, satellite observations are limited to the surface, as they capture only surface-level radiation signals~\cite{li2023satellite}. What truly drives human comfort, ecosystem dynamics, and urban planning are the physical properties that describe phenomena occurring above the ground, such as the near-surface air temperature (NSAT) measured 2 meters above the surface~\cite{chen2022high, yang2022modulation, fan2024exploring}. This mismatch between what satellites observe and what physically affects humans creates a gap in how we monitor the Earth~\cite{sun2020trend}, as shown in \cref{fig:problem_illustration}.

\vspace{1pt}

Near-ground sensors provide precise and frequent NSAT measurements, but their spatial coverage is sparse and uneven~\cite{cao2013instrumental, du2022novel, wang2025novel}. Consequently, they fail to capture fine-scale temperature variability across heterogeneous environments~\cite{xu2018mapping}. This sparsity can bias downstream applications such as climate modeling, epidemiology, and urban heat assessment~\cite{kloog2015using, schwingshackl2024high, tarek2021uncertainty}. Leveraging satellite-derived surface properties as proxies for NSAT has therefore gained traction, given their strong physical coupling through surface-atmosphere energy exchange and their continuous spatial coverage~\cite{xu2018mapping}. Among existing approaches, physical models~\cite{pape2004modelling, sun2005air} and deep learning (DL)~\cite{shen2020deep, dai2024urban, lee2025estimating} methods have shown remarkable potential for NSAT estimation. Physical models estimate NSAT from surface energy balance equations but depend on variables often unavailable from satellite data~\cite{sun2005air}. In contrast, DL methods learn nonlinear mappings between satellite surface features, auxiliary variables, and NSAT without explicit physical modeling.

\vspace{1pt}

Physics-Informed Neural Networks (PINNs) bridge data-driven learning and physical modeling by embedding partial differential equations (PDEs) directly into the training process~\cite{luo2025physics}. They have shown promise in both computer vision (CV)~\cite{zhang2021physics, gong2024physics, ji2025pomp} and remote sensing (RS)~\cite{liu2022physics, shi2024physics, costa2025dani} tasks where physical laws govern observed phenomena. Existing PINNs are either supervised~\cite{raissi2019physics, muller2023deep}, relying on labeled data, or self-supervised~\cite{yaman2020self, yan2023st}, driven solely by physics-based losses. However, semi-supervised PINNs remain largely unexplored, despite their potential to leverage both labeled and unlabeled pixels by exploiting spatial and physical neighborhood information.

\vspace{1pt}
In this paper, we take a step toward semi-supervised PINNs for sparse measurements. We propose SPyCer, a semi-supervised physics-guided network designed for continuous NSAT estimation by proxy from satellite imagery. Our key insight is that, under sparse supervision, discarding neighboring pixels, even if unlabeled, overlooks valuable physical information, as these regions encode thermodynamic interactions that affect the observed NSAT. Therefore, each near-ground sensor is projected onto the satellite grid to form a local patch centered on its corresponding pixel. To guide the training, SPyCer employs a semi-supervised strategy. Supervision is applied at the central pixel using ground-truth NSAT and physics-informed constraints derived from the surface energy balance (SEB) and the advection-diffusion-reaction (ADR) PDEs, while predictions for surrounding pixels contribute only through physics-derived constraints. The contextual influence of neighboring pixels is modeled using a multi-head convolutional attention mechanism guided by land cover spectral indices, including the Normalized Difference Vegetation Index (NDVI), Normalized Difference Water Index (NDWI), and Normalized Difference Built-up Index (NDBI). Additionally, a Gaussian distance modulation is employed to reinforce spatial coherence. By coupling sparse supervision with dense physics-based learning, SPyCer produces continuous, physically consistent NSAT estimates from satellite imagery. Our key contributions are as follows:

\begin{itemize}
    \item We propose a novel semi-supervised PINN designed for sparse measurements. It unifies labeled sparse near-ground sensor measurements and unlabeled continuous satellite pixels within a single training framework.

    \item We embed physical constraints from the SEB and the ADR PDEs directly into the learning objective, ensuring the network produces physically consistent predictions.

    \item We introduce a contextual convolutional spatial attention mechanism that determines the physical influence of neighboring pixels on the central labeled measurement.
\end{itemize}

    




\section{Related Works}

\begin{figure*}[ht]
  \centering
  \includegraphics[width=0.95\textwidth]
  {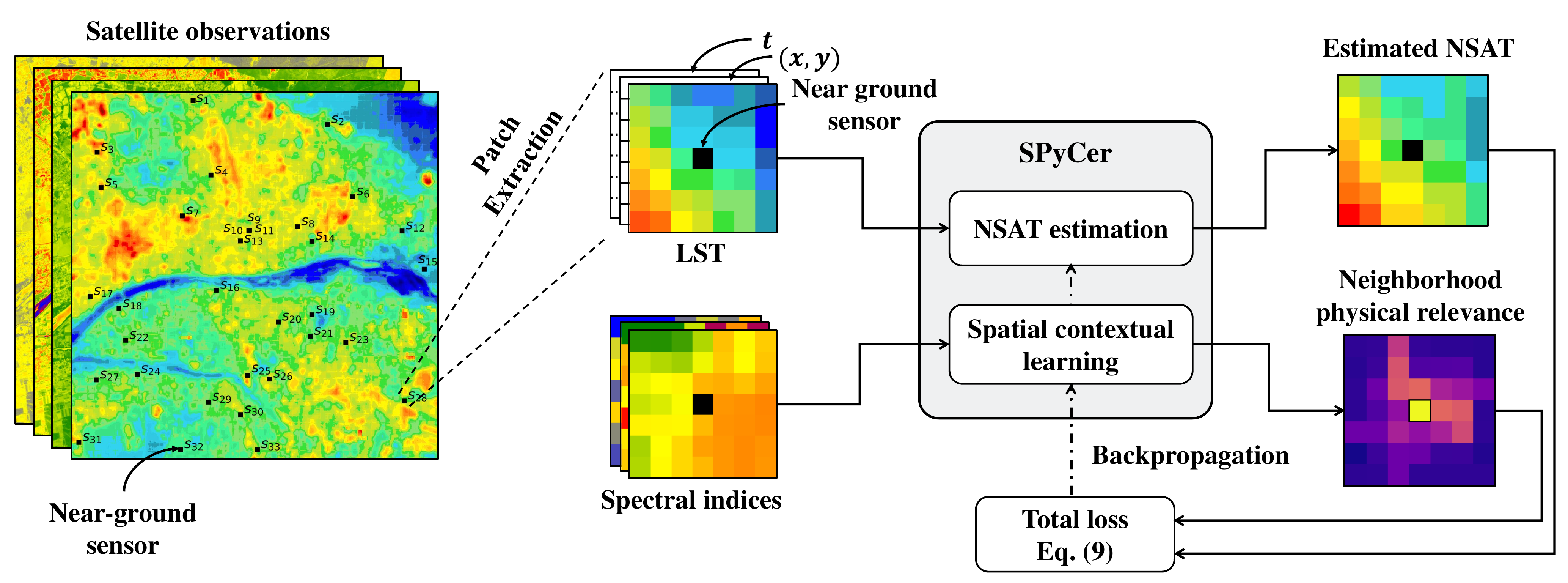}
    \caption{Overview of SPyCer. For each near-ground sensor, a square patch of satellite-derived LST and auxiliary variables is extracted, representing the local spatial neighborhood. SPyCer predicts the NSAT at the central pixel while leveraging contextual information from neighboring pixels through a physics-informed semi-supervised strategy. Learnable spatial contextual weights quantify the physical relevance of each neighbor, which guide the total loss defined in \cref{eq:totalloss_patch}.}
    
  \label{fig:spycer_main}
\end{figure*}

Our work builds upon research directions aimed at learning from sparse measurements, capturing physical constraints, and leveraging spatial context for improved predictions.

\smallskip

\noindent \textbf{Deep Learning for Sparse Measurement} deals with learning from datasets where only a limited portion of the data is labeled or measured~\cite{chai2020deep, cuypers2021deep, ke2021universal, hu2022deep}. This is widely used in environmental science, where measurements often exist only at irregularly distributed locations~\cite{wang2020weakly, tseng2021learning, moraes2025weakly}.  Traditional DL networks struggle to generalize under such limited supervision. Several approaches have attempted to address this challenge by spatial interpolation of sparse measurements~\cite{kirkwood2022bayesian, cutolo2024cloinet, archambault2023multimodal}, by designing graph-based models to propagate information across unevenly spaced data~\cite{li2023rainfall, lou2024non}, or by employing semi-supervised, weakly-supervised, and self-supervised frameworks~\cite{tian2021early, moraes2025weakly}. Nonetheless, these strategies often fail to account for the underlying physical dependencies that govern spatial variability.
\smallskip

\noindent \textbf{Physics-Informed Neural Networks} embed physical knowledge into DL models by enforcing that predictions respect governing equations, typically in the form of PDEs ~\cite{cuomo2022scientific, luo2025physics}. A common approach defines a residual by setting the PDE to zero and incorporating it into the loss function to be minimized during training~\cite{grossmann2024can, luo2025physics}. PINNs have been widely applied to heat diffusion, advection, and fluid dynamics problems~\cite{ cai2021physics, liu2022deep, gruszczynski2025beyond}, and more recently to RS tasks where spatial and temporal dynamics are governed by known physical laws~\cite{liu2022physics, shi2024physics, costa2025dani}. Existing PINNs frameworks follow either a supervised approach~\cite{raissi2019physics, muller2023deep}, where training relies on available labeled data alongside enforcing physical consistency, or a self‑supervised approach~\cite{yaman2020self, yan2023st}, in which the network is guided solely by the physical loss without using labeled observations. Semi-supervised PINNs, which can leverage both sparse labeled points and dense neighboring information, remain relatively unexplored, despite their potential for CV and RS scenarios.


\smallskip

\noindent \textbf{Spatial Contextual Learning} focuses on modeling spatial dependencies among neighboring pixels to improve local predictions and structural coherence~\cite{liu2018picanet, hassani2023neighborhood}. Early convolutional networks captured local context via fixed receptive fields~\cite{long2015fully}, but treated all neighbors equally within the kernel, failing to account for the varying relevance of each pixel. Recent methods apply attention mechanisms to adaptively weight neighboring pixels based on learned relevance~\cite{ liu2018picanet, li2018pyramid, huang2019ccnet, hassani2023neighborhood}, and leverage graph-based reasoning to propagate information across irregular neighborhoods~\cite{velickovic2017graph, xu2019spatial, kavran2023graph}. Incorporating spatial contextual learning into PINNs remains critical, as it can capture strong physical interactions and improve generalization, particularly under sparse supervision.


\section{Method}

We start with an overview of SPyCer in \cref{sec:overview} and describe the physical foundations in \cref{sec:physics_foundations}. Then, we present the network architecture in \cref{sec:network_architecture} and the semi-supervised physics-guided loss in \cref{sec:SS_PINN_loss}.

\subsection{Overview}
\label{sec:overview}

Let $\mathcal{S} = \{s_i\}_{i=1}^{N}$ denote a sparse network of $N$ near-ground sensors, where each sensor $s_i$ measures NSAT and is defined as $s_i = (x_i, y_i, t_i, T_a^i)$. Here, $(x_i, y_i)$ are the coordinates of the sensor after projection onto the satellite grid, $t_i$ is the observation time, and $T_a^i$ is the recorded NSAT. Our goal is to estimate a continuous NSAT field across space, $T_a(x, y, t),  \forall (x, y) \in \Omega$, using satellite imagery and limited near-ground measurements.


\smallskip

Each near-ground sensor defines the center of a local satellite image patch that represents the spatial context of the surrounding neighborhood and includes both primary and auxiliary inputs. The primary input is the 10-meter satellite-derived LST ($T_s$), while auxiliary inputs ($\mathbf{X}_{\text{aux}}$) provide spatial, temporal, and contextual cues. Spatial coordinates $(x, y)$ are represented in the Universal Transverse Mercator (UTM) projection, allowing distances to be expressed in meters, which is essential for computing PDEs. Temporal information $t$ is encoded using the day of year through sine and cosine functions, $\sin\!\big(2 \pi t / 365\big)$ and $\cos\!\big(2 \pi t / 365\big)$, to capture seasonal variations in temperature patterns~\cite{verma2024climode}. Contextual inputs are derived from monthly aggregated spectral indices that describe land cover characteristics, including the NDVI, NDWI, and NDBI.



\vspace{1pt}
Formally, we aim to learn a function $F_\theta$, parameterized by $\phi$, that maps these inputs to NSAT, as defined in \cref{eq:F}.
\begin{equation}
    T_a(x,y,t) \approx F_\theta\big(T_s(\mathcal{P}_{x,y}), \mathbf{X}_{\text{aux}}(\mathcal{P}_{x,y})\big),
\label{eq:F}
\end{equation}

\noindent where $\mathcal{P}_{x,y}$ denotes the patch of satellite pixels centered at location $(x,y)$. Within each patch, SpyCer is trained using the ground-truth NSAT at the central pixel along with the physics-informed loss, while predictions for surrounding pixels are constrained solely through physics-based regularization. SPyCer further employs learnable attention weights guided by land cover information and surface characteristics (NDVI, NDWI, NDBI) to quantify the relative physical influence of neighboring pixels. The main methodology is summarized in \cref{fig:spycer_main}, and its main components will be detailed in the following subsections.

\subsection{Physical Foundations}
\label{sec:physics_foundations}

\begin{figure}[ht]
  \centering
  \includegraphics[width=0.45\textwidth]
  {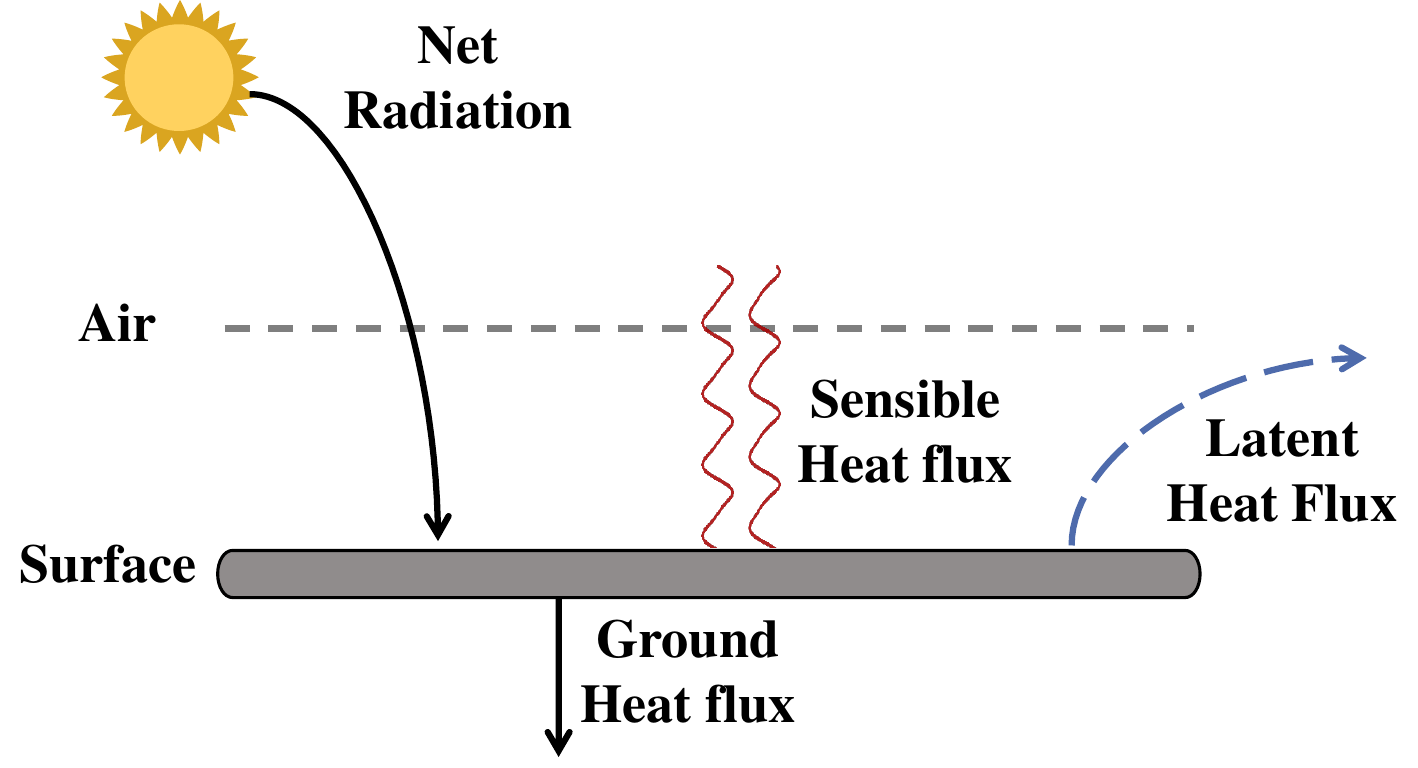}
  \caption{Representation of the SEB. Net radiation, primarily from solar and atmospheric inputs, is partitioned into three flux components: sensible heat flux ($H$) transferring energy to the air, latent heat flux ($LE$) driving evapotranspiration, and ground heat flux ($G$) conducted into the soil.}

  \label{fig:SEB}
\end{figure}

\textbf{Surface energy balance} (SEB) governs how radiative energy at the Earth's surface is partitioned into different fluxes that heat the air, ground, and water. It can be expressed as in \cref{eq:SEB}~\cite{foken2024microclimatology, su2002surface} and illustrated in \cref{fig:SEB}.
\begin{equation}
    R_n = H + LE + G
\label{eq:SEB}
\end{equation}
\noindent where \(R_n\) is the net radiation, \(H\) the sensible heat flux, \(LE\) the latent heat flux, and \(G\) the ground heat flux. 

The sensible heat flux, which directly links LST and NSAT, is given by \cref{eq:H}~\cite{lagouarde1992daily}.
\begin{equation}
    H = \rho c_p \frac{T_s(x,y,t) - T_a(x,y,t)}{r_a}
\label{eq:H}
\end{equation}
\noindent where $T_s(x,y,t)$ and $T_a(x,y,t)$ are the LST and NSAT, respectively, at spatial location $(x,y)$ and time $t$, and $\rho$, $c_p$, and $r_a$ denote the air density, specific heat capacity of air, and aerodynamic resistance.  This formulation establishes the physical coupling between LST and NSAT, which forms the foundation of our physics-based model.


\vspace{1pt}
\noindent \textbf{Advection-Diffusion-Reaction} (ADR) models the spatio-temporal evolution of a scalar quantity $\phi(x,y,t)$, under the combined effects of transport, mixing, and local sources or sinks~\cite{eliasof2023adr, union2013advection, clairambault2013reaction}. In our settings, $\phi$ represents the NSAT. It can be expressed as in \cref{eq:ADR}.
\begin{equation}
\frac{\partial T_a}{\partial t} = 
\underbrace{- \mathbf{u} \cdot \nabla T_a}_{\text{Advection}} +
\underbrace{\nabla \cdot (K \nabla T_a)}_{\text{Diffusion}} +
\underbrace{f(T_a, \theta_r)}_{\text{Reaction}},
\label{eq:ADR}
\end{equation}
\noindent where the advection term models heat transport by air motion, with $\mathbf{u}$ as the local wind velocity. The diffusion term represents turbulent mixing, with $K$ as the diffusion coefficient. The reaction term is a local function accounting for sources, sinks, or interactions of NSAT with its environment, parameterized by $\eta_r$. In our case, the dominant local source of heat is the energy flux transmitted from the land surface to the air. We approximate this by the sensible heat flux ($H$) from the SEB equation, $f(T_a, \eta_r) \approx \alpha (T_s(x,y,t) - T_a(x,y,t))$, where $\alpha = \rho c_p / r_a$. 


\vspace{1pt}
For computational tractability, we assume negligible advection ($\nabla T_a \approx 0$°C/m) and a fixed reference height (2 meters), reducing the ADR equation to a two-dimensional diffusion-reaction system as defined in \cref{eq:ADE_simplified}.
\begin{equation}
\frac{\partial T_a}{\partial t} = 
K \left( \frac{\partial^2 T_a}{\partial x^2} + \frac{\partial^2 T_a}{\partial y^2} \right)
+ \alpha (T_s(x,y,t) - T_a(x,y,t)),
\label{eq:ADE_simplified}
\end{equation}
\noindent This assumption is justified by the high spatial resolution of satellite observations (10 meters) and the limited spatial extent of each local patch. At this fine scale, horizontal heat advection is negligible compared to diffusive mixing and surface-atmosphere energy exchange.

\vspace{1pt}
Additionally, since temporal information is encoded as $\mathbf{t} \mapsto \big(\sin(2 \pi t / 365), \cos(2 \pi t / 365)\big)$, using the chain rule for one independent variable, the derivative of NSAT with respect to \(t\) in \cref{eq:ADE_simplified} can be expressed as in \cref{eq:diff_t}.
\begin{equation}
\frac{\partial T_a}{\partial t} = \frac{2 \pi}{365} 
\Bigg(
\frac{\partial T_a}{\partial \sin} \cdot \cos\Big(\frac{2 \pi t}{365}\Big) 
- \frac{\partial T_a}{\partial \cos} \cdot \sin\Big(\frac{2 \pi t}{365}\Big)
\Bigg),
\label{eq:diff_t}
\end{equation}

\subsection{Network Architecture}
\label{sec:network_architecture}
SPyCer consists of two main components: the NSAT estimation network and the spatial contextual learning module, which are detailed below.


\smallskip

\noindent \textbf{NSAT Estimation via Local Patches.} The NSAT estimation network in SPyCer is a lightweight ResNet-style CNN~\cite{he2016deep} that predicts NSAT from 10 meters resolution satellite image patches along with auxiliary spatial and temporal characteristics. Residual connections within the CNN enable efficient feature extraction and stabilize training by improving gradient flow~\cite{huang2023revisiting}, which is particularly beneficial under sparse supervision. We adopt a ResNet-style architecture because the physical neighborhood influencing NSAT is inherently localized, which keeps the receptive field limited. This allows the network to capture relevant patterns effectively without the computational overhead of more complex architectures

\vspace{1pt}
Each input corresponds to a local satellite image patch centered on a near-ground sensor, whose measurement is used solely for supervision in the loss function. During inference, no information from the sensor is required. In our experiments, the patch size was fixed to $7 \times 7$ pixels, corresponding to a $70 \times 70$~m area on the ground. It was selected via grid search to balance capturing the dominant spatial footprint of surface-atmosphere interactions while minimizing redundant context.




\smallskip
\noindent \textbf{Spatial Contextual Learning.} To model the physical influence from neighboring pixels, SPyCer employs a spatial contextual learning strategy that represents local interactions within each satellite image patch. The objective is to quantify the contribution of each neighbor $(x',y') \in \mathcal{P}_{x,y} \setminus \{(x,y)\}$ to the central NSAT estimation $({x,y})$, through learnable weights. 

\vspace{1pt}
We employ a multi-head convolutional attention mechanism to generate attention maps for neighboring pixels, where each head consists of convolutional layers that produce a spatial weighting based on land cover characteristics. Dropout with a rate of $0.15$ is applied within each head to prevent overfitting during training. Specifically, the attention is guided by the spectral indices ($\mathbf{X}_{\text{Ind}}$), NDVI, NDBI, and NDWI, extracted from Sentinel-2 imagery at 10-meter. Due to the 5-day revisit cycle and frequent cloud cover, we use one representative aggregated composite per month for each index. This is justified as these indices vary slowly over a monthly period and serve only as land cover indicators. To avoid bias toward the central pixel, its contribution is explicitly excluded when computing attention. The attention weights are defined as in \cref{eq:attn}.
\begin{equation}
w_{x',y'} = \frac{1}{H} \sum_{h=1}^{H} 
\mathrm{softmax}\Big(g_h(\mathbf{X}_{\text{Ind}}(x',y'))\Big),
\label{eq:attn}
\end{equation}
\noindent where $H$ is the number of attention heads and $g_h$ denotes a lightweight CNN that maps the spectral features to a scalar relevance score for each head.



\vspace{1pt}
Since NSAT is strongly influenced by spatial proximity, we modulate the learned attention weights with a Gaussian distance kernel, as expressed in \cref{eq:gaussian}. This reinforces local spatial coherence and helps the multi-head convolutional attention converge faster and more stably.
\begin{equation}
\hat{w}_{x,y} = 
w_{x,y} \cdot 
\mathrm{e}^{-\frac{x^2 + y^2}{2\sigma^2}},
\label{eq:gaussian}
\end{equation}
\noindent where $\sigma$ controls the spatial decay and is fixed to $1.5$ in our experiments. The resulting weights are normalized over the patch to form a probability distribution of neighbor influence, $\sum_{(x',y') \in \mathcal{P}_{x,y} \setminus \{(x,y)\}} 
\tilde{w}_{x',y'} = 1$.

\subsection{Semi-Supervised Physics-Guided Loss}
\label{sec:SS_PINN_loss}
SPyCer adopts a semi-supervised training strategy that combines sparse near-ground measurements with dense supervision derived from physical laws. This enables SPyCer to learn physically consistent NSAT estimations even in regions without near-ground observations. Precisely, for each satellite image patch, the model is trained using the observed NSAT at the central pixel together with a physics-informed loss, while predictions for neighboring pixels are guided solely by physics-based regularization weighted by their learned contextual relevance. The total patch loss is defined as in \cref{eq:totalloss_patch}.
\begin{align}
\mathcal{L}_{\text{patch}} =\;& 
\underbrace{\mathcal{L}_{\text{sup}}\!\big(\hat{T}_a(x,y), T_a^{\text{center}}\big)}_{\text{central supervised loss}} \;+\;
\lambda \Bigg(
\underbrace{\mathcal{L}_{\text{phys}}\!\big(\hat{T}_a(x,y)\big)}_{\text{central physics loss}}  \nonumber \\
& + \sum_{(x',y') \in \mathcal{P}_{x,y} \setminus \{(x,y)\}} 
\hat{w}_{x',y'} \,
\underbrace{\mathcal{L}_{\text{phys}}\!\big(\hat{T}_a(x',y')\big)}_{\text{neighbor physics loss}}
\Bigg)
\label{eq:totalloss_patch}
\end{align}
\noindent where $T_a^{\text{center}}$ is the observed NSAT at the central pixel, $\hat{w}_{x',y'}$ are the contextual learned weights defined in \cref{{eq:gaussian}}, and $\lambda$ controls the relative influence of the physics-based supervision. In our settings, we fixed $\lambda = 0.9$. The physics loss $\mathcal{L}_{\text{phys}}$ enforces local compliance with the simplified ADR constraint in \cref{eq:ADE_simplified}. Here, we minimize the residual defined in \cref{eq:Lphys_patch}.
\begin{equation}
\begin{aligned}
\mathcal{L}_{\text{phys}}\big(\hat{T}_a(x',y')\big) =
\Big\|
\frac{\partial \hat{T}_a(x',y')}{\partial t}
- K \nabla^2 \hat{T}_a(x',y') \\
- \alpha \big(T_s(x',y') - \hat{T}_a(x',y')\big)
\Big\|_2^2
\end{aligned}
\label{eq:Lphys_patch}
\end{equation}
\noindent where $\alpha$ and $K$ are hyperparameters. In SPyCer, the best results were obtained with $\alpha = 0.5$ and $K = 0.8$.


\section{Experiments}

\begin{table*}[htbp]
\centering
\caption{Quantitative comparison of SPyCer against baseline methods (LR~\cite{yang2017evaluation, liu2017evaluating, wang2017comparison}, RF~\cite{noi2017comparison, zhang2024remotely, karagiannidis2025real}, GB~\cite{vedri2025empirical, limoncella2025machine}, MLP~\cite{shen2020deep}). Metrics (RMSE and MAE) are reported per month over 100 Monte Carlo runs. Values indicate mean ± standard deviation. The blue, yellow, and orange colors represent the top three results in order.}

\renewcommand{\arraystretch}{1.1}
\begin{tabular}{c|c@{\hskip 9pt}c@{\hskip 9pt}c@{\hskip 9pt}c@{\hskip 9pt}c@{\hskip 9pt}c@{\hskip 9pt}c|c}
\toprule
\textbf{Method} & \textbf{Metric} & \textbf{April} & \textbf{May} & \textbf{June} & \textbf{July} & \textbf{August} & \textbf{September} & \textbf{Average} \\
\midrule

\multirow{2}{*}{LR} 
& RMSE & 3.66 $\pm$0.67 & 4.09 $\pm$0.47 & 3.56 $\pm$0.62 & 3.89 $\pm$0.64 & \cellcolor{thirdbest}{3.61 $\pm$0.69} & 3.91 $\pm$0.66 & 3.77 $\pm$0.54  \\
& MAE  & 2.95 $\pm$0.56 & 3.39 $\pm$0.42 & 2.99 $\pm$0.59 & 3.26 $\pm$0.55 & \cellcolor{thirdbest}{3.01 $\pm$0.60} & 3.30 $\pm$0.66 & 3.11 $\pm$0.49 \\

\noalign{\vskip 2pt}
\multirow{2}{*}{RF} 
& RMSE & \cellcolor{thirdbest}{2.91 $\pm$0.60} & \cellcolor{secondbest}{2.92 $\pm$0.51} & 3.27 $\pm$0.67 & 3.52 $\pm$0.64 & 3.81 $\pm$0.80 & \cellcolor{thirdbest}{3.70 $\pm$0.75} & 3.33 $\pm$0.54 \\
& MAE & \cellcolor{thirdbest}{2.36 $\pm$0.55} & \cellcolor{secondbest}{2.41 $\pm$0.49} & 2.67 $\pm$0.59 & 2.96 $\pm$0.60 & 3.18 $\pm$0.73 & 3.17 $\pm$0.71 & 2.73 $\pm$0.49\\

\noalign{\vskip 2pt}
\multirow{2}{*}{GB} 
& RMSE & 2.93 $\pm$0.65 & \cellcolor{thirdbest}{2.99 $\pm$0.68} & \cellcolor{thirdbest}{3.19 $\pm$0.72} & \cellcolor{thirdbest}{3.45 $\pm$0.72} & 3.69 $\pm$0.90 & \cellcolor{secondbest}{3.65 $\pm$0.77} & \cellcolor{thirdbest}{3.29 $\pm$0.64}  \\
& MAE & 2.38 $\pm$0.60 & \cellcolor{thirdbest}{2.47 $\pm$0.58} & \cellcolor{thirdbest}{2.62 $\pm$0.63} & \cellcolor{thirdbest}{2.90 $\pm$0.65} & 3.08 $\pm$0.77 & \cellcolor{thirdbest}{3.14 $\pm$0.73} & \cellcolor{thirdbest}{2.70 $\pm$0.56} \\

\noalign{\vskip 2pt}
\multirow{2}{*}{MLP} 
& RMSE & \cellcolor{secondbest}{2.86 $\pm$0.54} & 3.25 $\pm$0.40 & \cellcolor{secondbest}{2.77 $\pm$0.58} & \cellcolor{secondbest}{3.10 $\pm$0.55} & \cellcolor{secondbest}{2.78 $\pm$0.67} & \cellcolor{thirdbest}{3.70 $\pm$0.95} & \cellcolor{secondbest}{3.03 $\pm$0.46} \\
& MAE & \cellcolor{secondbest}{2.33 $\pm$0.43} & 2.66 $\pm$0.33 & \cellcolor{secondbest}{2.24 $\pm$0.48} & \cellcolor{secondbest}{2.55 $\pm$0.48} & \cellcolor{secondbest}{2.23 $\pm$0.55} & \cellcolor{secondbest}{3.05 $\pm$0.81} & \cellcolor{secondbest}{2.43 $\pm$0.37}  \\

\hline 
\noalign{\vskip 2pt}
\multirow{2}{*}{SPyCer} 
& RMSE & \cellcolor{myblue}2.27 $\pm$ 0.21 & \cellcolor{myblue}2.46 $\pm$ 0.20 & \cellcolor{myblue}2.01 $\pm$ 0.24 & \cellcolor{myblue}2.46 $\pm$ 0.26 & \cellcolor{myblue}2.04 $\pm$ 0.22 & \cellcolor{myblue}2.43 $\pm$ 0.35 & \cellcolor{myblue}2.27 $\pm$ 0.07 \\
& MAE  & \cellcolor{myblue}1.85 $\pm$ 0.19 & \cellcolor{myblue}1.99 $\pm$ 0.19 & \cellcolor{myblue}1.60 $\pm$ 0.19 & \cellcolor{myblue}2.05 $\pm$ 0.24 & \cellcolor{myblue}1.65 $\pm$ 0.19 & \cellcolor{myblue}1.99 $\pm$ 0.29 & \cellcolor{myblue}1.83 $\pm$ 0.07 \\

\bottomrule
\end{tabular}
\label{tab:RF_monthly_metrics}
\end{table*}

\subsection{Experimental setup}
\textbf{Dataset.} We evaluate SPyCer on a regional dataset covering an urban and peri-urban area in north-central France with mixed land cover and a temperate climate. The dataset contains $59$ cloud-free LST at 10-meter resolution acquired between 1st April 2025 and 30th September 2025, of size $1200 \times 1200$. LST images are generated through spatio-temporal fusion of Terra MODIS, Landsat 8, and Sentinel-2 satellites, with a reported average error of 1.95$^{\circ}$C~\cite{bouaziz2025wgast}. It also includes NDVI, NDBI, and NDWI at the same resolution. In this area, 33 near-ground sensors coexist, measuring NSAT and forming a sparse network of measurements.


\smallskip

\noindent \textbf{Evaluation Strategy.} We randomly split the available near-ground sensors into 80\% for training and 20\% for testing over the six-month evaluation period. SPyCer predicts NSAT at the test sensor locations, and the predicted values are compared against the corresponding ground-truth NSAT measurements. To ensure robustness and validate spatial generalization, this random split is repeated $100$ times, each with a different selection of training and test sensors, following a Monte Carlo cross-validation protocol.


\smallskip

\noindent \textbf{Implementation Details.} SPyCer is implemented in PyTorch. Training is performed for $2000$ epochs using a learning rate of $3\times10^{-3}$ for the NSAT estimation model and $5\times10^{-5}$ for the attention module. First and second order derivatives required for the physics-based loss in \cref{eq:Lphys_patch} are computed using PyTorch automatic differentiation. All experiments are conducted on a physical server equipped with an NVIDIA RTX A6000 GPU. The open-source code will be released upon acceptance.


\smallskip

\noindent \textbf{Baslines.} We compare SPyCer against state-of-the-art baselines commonly used for NSAT estimation: Linear Regression (LR)~\cite{yang2017evaluation, liu2017evaluating, wang2017comparison}, Random Forest (RF)~\cite{noi2017comparison, zhang2024remotely, karagiannidis2025real} with $300$ estimators and maximum depth $15$, Gradient Boosting (GB)~\cite{vedri2025empirical, limoncella2025machine} with $500$ estimators, max depth $6$, and learning rate $0.05$, and a Multi-Layer Perceptron (MLP)~\cite{shen2020deep} with a learning rate $0.03$. These baselines perform direct pixel-wise supervision. To the best of our knowledge, no existing method leverages neighboring pixels in sparse measurement networks for NSAT estimation.

\subsection{Quantitative Assessment}
We evaluate SPyCer and the baselines using Root Mean Square Error (RMSE) and Mean Absolute Error (MAE) on test sensor locations across the six-month period. \cref{tab:RF_monthly_metrics} reports the mean and standard deviation over 100 Monte Carlo folds. SPyCer consistently outperforms all baselines across all months. Notably, it produces stable results even during months with higher variability. For instance, in August 2025, MLP RMSE increases from $2.78$ in August to $3.70$ in September, whereas SPyCer only varies from $2.04$ to $2.4$3. A similar trend is observed for MAE, where MLP fluctuates from $2.23$ to $3.05$, while SPyCer remains in the narrow range $1.65-1.99$. On average, SPyCer reduces RMSE by $25.08\%$ compared to MLP, $31.00\%$ compared to GB, $31.83\%$ compared to RF, and $39.79\%$ compared to LR. Similarly, average MAE decreases by $24.70\%$ relative to MLP, $32.22\%$ to GB, $33.97\%$ to RF, and $41.16\%$ to LR. Moreover, the standard deviation of SPyCer remains relatively low across the 100 random sensor selections, demonstrating robust generalization under varying test conditions.

\subsection{Qualitative Assessment}
\begin{figure*}[htbp]
  \centering
  \includegraphics[width=0.85\textwidth]
  {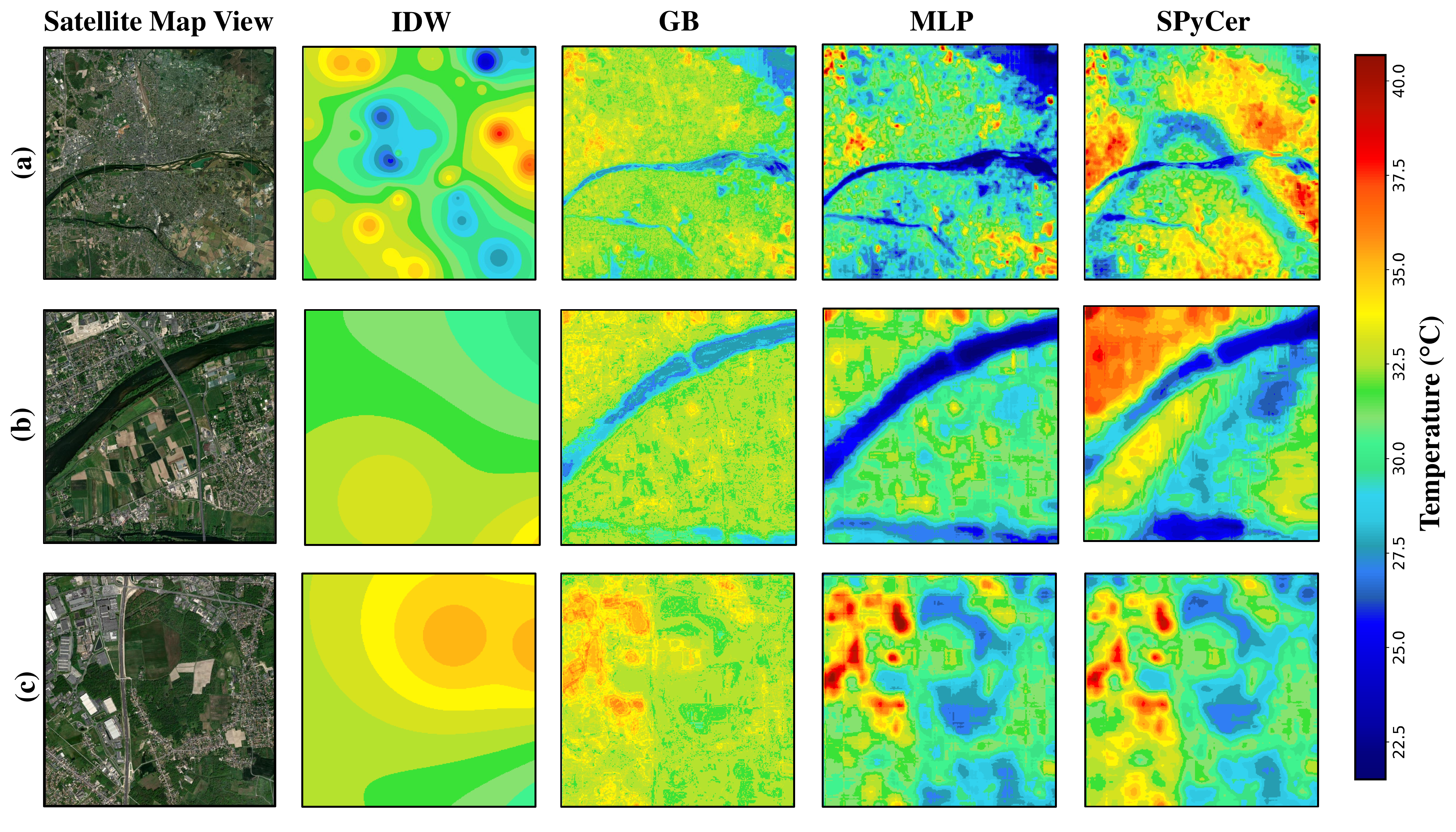}

    \caption{Qualitative comparison of NSAT estimates on 08 August 2025. Columns show IDW, GB, MLP, and SPyCer predictions, alongside the high-resolution satellite image for context. Rows correspond to (a) the full region, (b) a semi-urban corridor with a river, and (c) a major industrial zone. SPyCer accurately captures large-scale temperature patterns while resolving fine-scale variability, including cold regions, river features, and industrial hotspots, outperforming GB and MLP in spatial fidelity and local detail.}

  \label{fig:qualitative}
\end{figure*}

\cref{fig:qualitative} presents a qualitative comparison on 08 August 2025 between NSAT estimates from Inverse Distance Weighting (IDW), MLP and GB, the two best-performing baselines, and SPyCer, alongside the corresponding high-resolution satellite image. Three representative areas are highlighted: (a) the full region, (b) a semi-urban corridor intersected by a river, and (c) a major industrial zone, with (b) and (c) shown as zoomed-in areas. Other baselines are omitted due to inferior performance (\cref{tab:RF_monthly_metrics}). IDW, while lacking fine spatial detail, provides a reference for the general distribution of warm and cold regions by integrating all near-ground sensor measurements. Overall, GB produces uniform predictions that fail to capture local variability, whereas MLP introduces some finer structure but is noisy and inconsistent. In contrast, SPyCer captures complex spatial gradients, closely following the large-scale temperature patterns of IDW while enhancing local contrast and fine-scale variability. Specifically, the central cold region in \cref{fig:qualitative}.a is accurately reproduced by SPyCer but missed by MLP and GB. In \cref{fig:qualitative}.b, SPyCer differentiates the river, its banks, and the crossing bridge, whereas MLP oversmooths these features. In \cref{fig:qualitative}.c, both SPyCer and MLP successfully reconstruct the major industrial hotspot, while GB significantly underestimates it.

\subsection{Temporal Consistency Analysis}

\begin{figure}[htbp]
  \centering
  \includegraphics[width=0.45\textwidth]
  {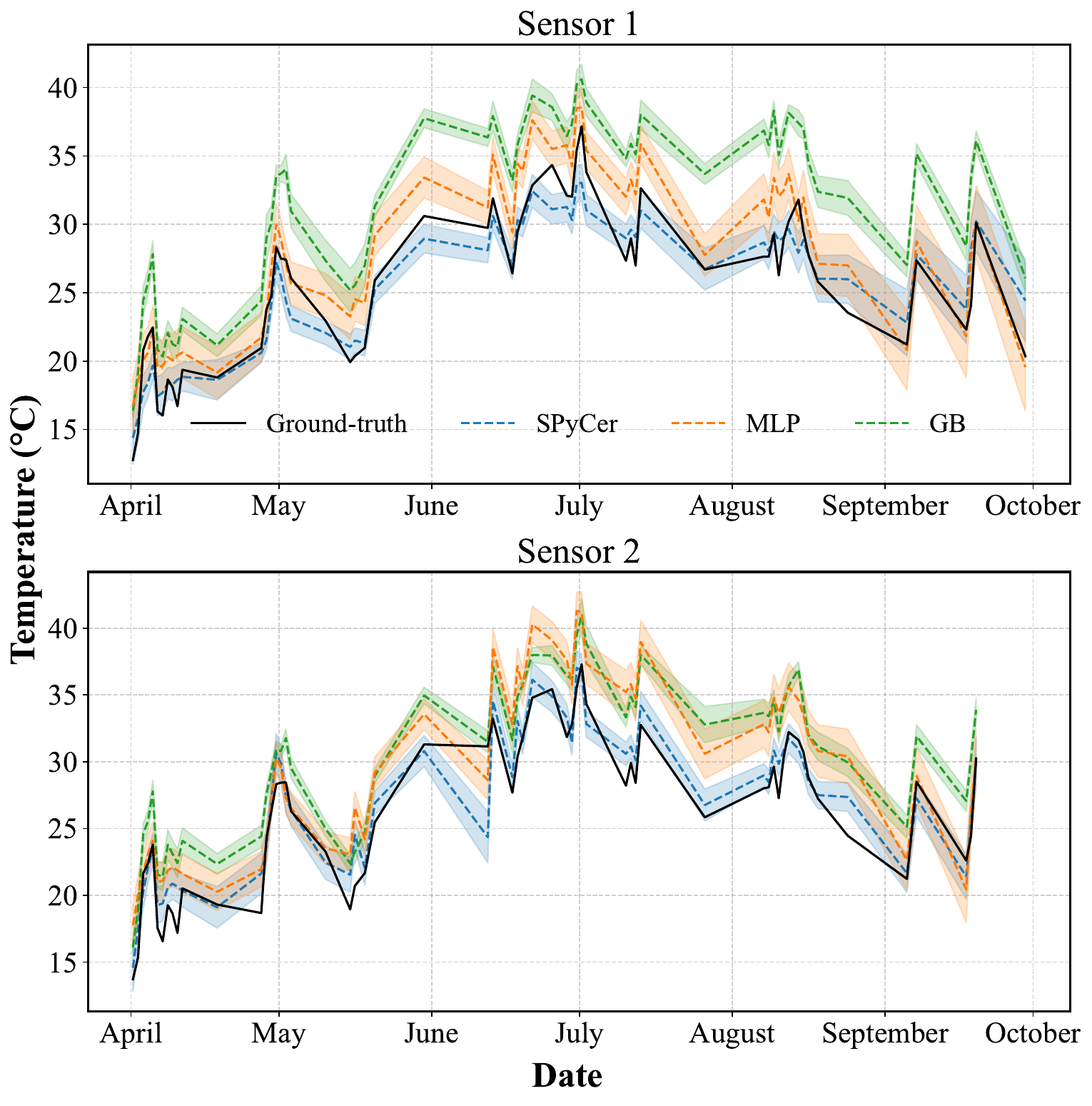}
    \caption{Temporal evolution of NSAT estimates from SPyCer, MLP, and GB compared to ground-truth measurements at two randomly selected near-ground sensors from April to October 2025. SPyCer closely follows both the amplitude and timing of temperature variations, whereas MLP and GB overestimate during warm periods and fail to capture fine-scale fluctuations.}

  \label{fig:temporal_curves}
\end{figure}

\cref{fig:temporal_curves} compares the temporal evolution of SPyCer’s estimations with those from MLP and GB against ground-truth measurements from two randomly selected near-ground sensors. The real observations exhibit pronounced short-term fluctuations and clear seasonal trends from April to October 2025. SPyCer’s estimations closely track the true NSAT dynamics for both sensors by accurately capturing both the amplitude and timing of temperature peaks. In contrast, GB and MLP tend to overestimate during warm periods and fail to reproduce local variations. Moreover, the standard deviation of SPyCer’s predictions remains consistently low throughout the evaluation.

\begin{figure*}[htbp]
  \centering
  \includegraphics[width=1\textwidth]
  {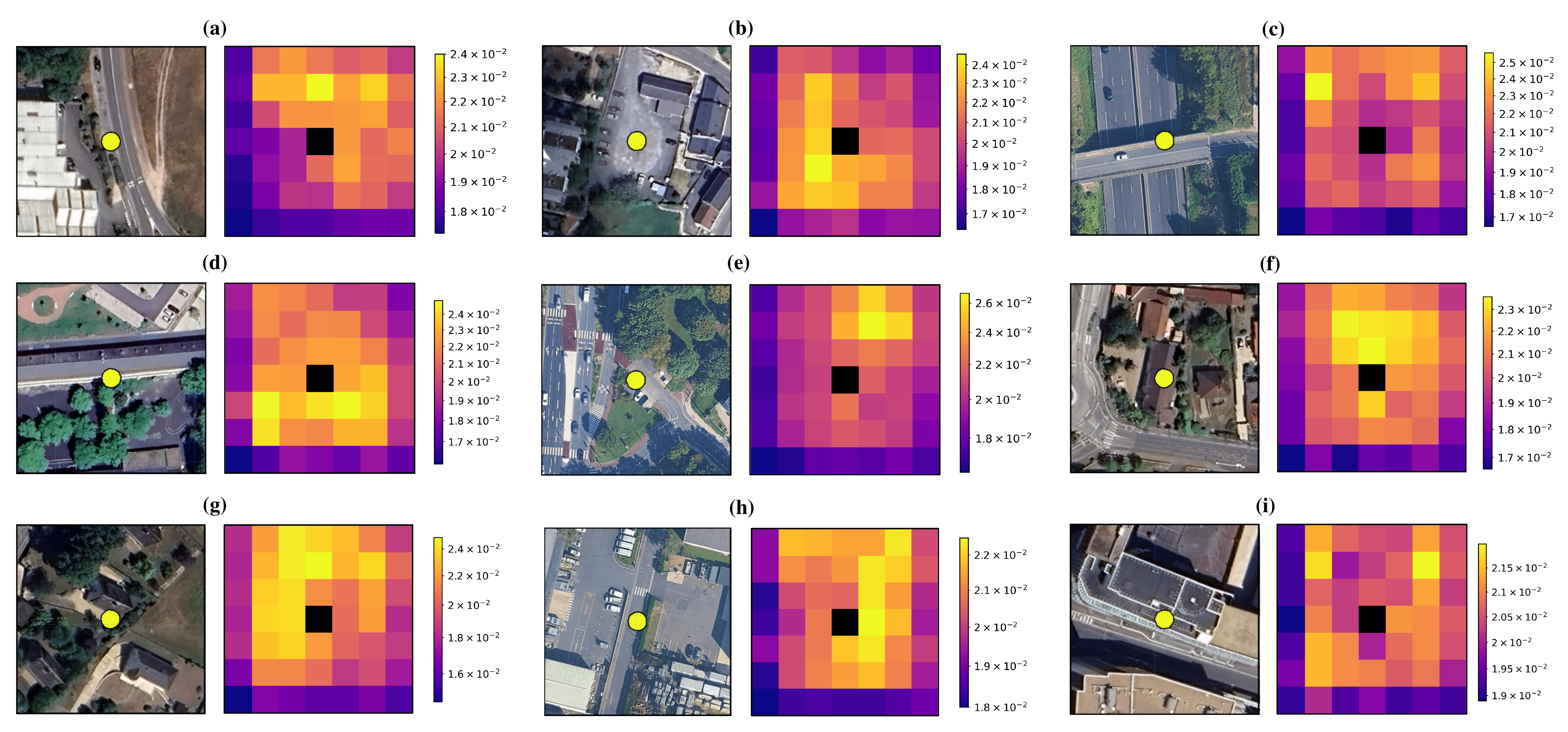}

  \caption{Spatial attention maps learned by SPyCer for nine randomly selected sensors. For each near-ground sensor, the left panel shows the local RGB satellite context (yellow dot marks the sensor location), and the right panel shows the corresponding $7\times7$ neighborhood attention map weights, where warmer colors indicate higher influence on the central NSAT prediction. SPyCer consistently emphasizes neighboring pixels with similar land-cover and thermal characteristics.}

  \label{fig:attention}
\end{figure*}

\subsection{Contextual Attention Analysis}

\cref{fig:attention} visualizes the contextual attention values maps learned by SPyCer for nine randomly selected sensors. Each pair of panels shows (left) the local RBG satellite image context centered on the sensor (yellow dot) and (right) the corresponding $7\times7$ attention map, where warmer colors indicate higher influence of neighboring pixels on the central prediction. Across all examples, SPyCer effectively adapts its attention patterns to the physical environment of each location.  In \cref{fig:attention}.a, the sensor is positioned near a road, with attention focused along the roadway and adjacent bare-soil areas, while nearby built-up structures receive notably less attention, consistent with their lower thermal similarity to the central pixel. In \cref{fig:attention}.b, positioned within a parking lot surrounded by concrete roofs, attention emphasizes impervious neighboring pixels rather than vegetated patches, effectively capturing the influence of surfaces with similar thermal properties and radiative behavior. Case \cref{fig:attention}.(c), located on a bridge, shows limited attention to the highway below, reflecting SPyCer’s ability to detect vertical offsets between surfaces, while highlighting vegetated edges at similar elevation. For \cref{fig:attention}.d, placed along a tree-lined urban walkway, attention peaks over vegetated zones, which indicates recognition of canopy cooling effects. In \cref{fig:attention}.e, situated at a small green intersection, the model focuses on the grassy and shaded northern area. In \cref{fig:attention}.f, within a residential block mixing roofs and gardens, attention concentrates on adjacent rooftops, linking similar built materials with comparable radiative behavior. For \cref{fig:attention}.g shows a suburban vegetated parcel where attention peaks over nearby grass and trees, confirming that SPyCer naturally clusters pixels with similar biophysical responses. In \cref{fig:attention}.h, from a semi-industrial parking zone, attention smoothly follows paved areas. Finally, \cref{fig:attention}.i illustrates a sensor near a building corner, where attention shifts toward the road and courtyard surfaces rather than the roof itself, associating thermal response with exposed impervious ground.

\subsection{Ablation Study}
To assess the contributions of the neighborhood physics and the Gaussian distance modulation in SPyCer, we conducted two ablation experiments. In Config.1, we removed all physics-based supervision from neighboring pixels. In Config.2, we retained neighborhood physics but removed the Gaussian distance kernel. Table \cref{tab:config_comparison} reports RMSE and MAE over 100 Monte Carlo folds for these configurations alongside the complete SPyCer model. Removing neighborhood physics (Config. 1) leads to a substantial drop in performance, which highlights the importance of incorporating neighboring pixels in the physics-guided loss. Eliminating the Gaussian kernel (Config. 2) also degrades accuracy. Overall, both components contribute to SPyCer’s superior accuracy and stability.

\begin{table}[ht]
\centering
\caption{Quantitative comparison of different configurations for NSAT estimation. Metrics (RMSE and MAE) are reported as mean ± standard deviation.}

\renewcommand{\arraystretch}{1.1}
\begin{tabular}{c|c c}
\toprule
\textbf{Configuration} & \textbf{RMSE} & \textbf{MAE} \\
\midrule
Config. 1 & 2.80 $\pm$ 0.28 & 2.26 $\pm$ 0.23 \\
Config. 2 & 2.57 $\pm$ 0.27 & 2.08 $\pm$ 0.23 \\

\noalign{\vskip 2pt}
\hline
\noalign{\vskip 2pt}
\cellcolor{myblue} SPyCer  & \cellcolor{myblue}2.27 $\pm$ 0.07 & \cellcolor{myblue}1.83 $\pm$ 0.07 \\
\bottomrule
\end{tabular}
\label{tab:config_comparison}
\end{table}

\section{Conclusion}

\noindent In this work, we propose SPyCer, a physics-informed semi-supervised framework for estimating continuous physically consistent NSAT from sparse near-ground measurements and satellite imagery. Each near-ground sensor is projected onto the satellite image grid, forming the center of a local satellite image patch. SPyCer leverages semi-supervised learning, using the ground-truth NSAT at the central pixel along with a physics-informed loss, while predictions for surrounding pixels are constrained solely through physics-based regularization. To quantify the physical influence of neighboring pixels, we employ a multi-head convolutional attention mechanism guided by land cover and surface characteristics. Experimental results demonstrate that SPyCer outperforms state-of-the-art methods in both quantitative and qualitative assessments, while maintaining temporal consistency, as further supported by attention map analysis of neighboring pixels.

\vspace{1pt}

Nonetheless, the current framework is limited by the fixed receptive field defined by the patch size, which constrains the spatial context considered for each NSAT estimation. Future work will explore multi-patch learning strategies to aggregate information across larger spatial extents.


{
    \small
    \bibliographystyle{ieeenat_fullname}
    \bibliography{main}

@article{ghamisi2025responsible,
  title={Responsible Artificial Intelligence for Earth Observation: Achievable and realistic paths to serve the collective good},
  author={Ghamisi, Pedram and Yu, Weikang and Marinoni, Andrea and Gevaert, Caroline M and Persello, Claudio and Selvakumaran, Sivasakthy and Girotto, Manuela and Horton, Benjamin P and Rufin, Philippe and Hostert, Patrick and others},
  journal={IEEE Geoscience and Remote Sensing Magazine},
  year={2025},
  publisher={IEEE}
}

@article{zhang2022artificial,
  title={Artificial intelligence for remote sensing data analysis: A review of challenges and opportunities},
  author={Zhang, Lefei and Zhang, Liangpei},
  journal={IEEE Geoscience and Remote Sensing Magazine},
  volume={10},
  number={2},
  pages={270--294},
  year={2022},
  publisher={IEEE}
}

@article{lu2025vision,
  title={Vision foundation models in remote sensing: A survey},
  author={Lu, Siqi and Guo, Junlin and Zimmer-Dauphinee, James R and Nieusma, Jordan M and Wang, Xiao and Wernke, Steven A and Huo, Yuankai and others},
  journal={IEEE Geoscience and Remote Sensing Magazine},
  year={2025},
  publisher={IEEE}
}

@article{li2013satellite,
  title={Satellite-derived land surface temperature: Current status and perspectives},
  author={Li, Zhao-Liang and Tang, Bo-Hui and Wu, Hua and Ren, Huazhong and Yan, Guangjian and Wan, Zhengming and Trigo, Isabel F and Sobrino, Jos{\'e} A},
  journal={Remote sensing of environment},
  volume={131},
  pages={14--37},
  year={2013},
  publisher={Elsevier}
}

@article{li2023satellite,
  title={Satellite remote sensing of global land surface temperature: Definition, methods, products, and applications},
  author={Li, Zhao-Liang and Wu, Hua and Duan, Si-Bo and Zhao, Wei and Ren, Huazhong and Liu, Xiangyang and Leng, Pei and Tang, Ronglin and Ye, Xin and Zhu, Jinshun and others},
  journal={Reviews of Geophysics},
  volume={61},
  number={1},
  year={2023},
  publisher={Wiley Online Library}
}

@article{chen2022high,
  title={A high-resolution monitoring approach of canopy urban heat island using a random forest model and multi-platform observations},
  author={Chen, Shihan and Yang, Yuanjian and Deng, Fei and Zhang, Yanhao and Liu, Duanyang and Liu, Chao and Gao, Zhiqiu},
  journal={Atmospheric Measurement Techniques},
  volume={15},
  number={3},
  pages={735--756},
  year={2022},
  publisher={Copernicus Publications G{\"o}ttingen, Germany}
}

@article{yang2022modulation,
  title={Modulation of wintertime canopy urban heat island (CUHI) intensity in Beijing by synoptic weather pattern in planetary boundary layer},
  author={Yang, Yuanjian and Guo, Min and Ren, Guoyu and Liu, Shuhong and Zong, Lian and Zhang, Yanhao and Zheng, Zuofang and Miao, Yucong and Zhang, Ying},
  journal={Journal of Geophysical Research: Atmospheres},
  volume={127},
  number={8},
  pages={e2021JD035988},
  year={2022},
  publisher={Wiley Online Library}
}

@article{fan2024exploring,
  title={Exploring the relationship between air temperature and urban morphology factors using machine learning under local climate zones},
  author={Fan, Chengliang and Zou, Binwei and Li, Jianjun and Wang, Mo and Liao, Yundan and Zhou, Xiaoqing},
  journal={Case Studies in Thermal Engineering},
  volume={55},
  pages={104151},
  year={2024},
  publisher={Elsevier}
}

@article{sun2020trend,
  title={The trend inconsistency between land surface temperature and near surface air temperature in assessing urban heat island effects},
  author={Sun, Tao and Sun, Ranhao and Chen, Liding},
  journal={Remote sensing},
  volume={12},
  number={8},
  pages={1271},
  year={2020},
  publisher={MDPI}
}

@article{cao2013instrumental,
  title={Instrumental temperature series in eastern and central China back to the nineteenth century},
  author={Cao, Lijuan and Zhao, Ping and Yan, Zhongwei and Jones, Phil and Zhu, Yani and Yu, Yu and Tang, Guoli},
  journal={Journal of Geophysical Research: Atmospheres},
  volume={118},
  number={15},
  pages={8197--8207},
  year={2013},
  publisher={Wiley Online Library}
}

@article{du2022novel,
  title={A novel fully coupled physical--statistical--deep learning method for retrieving near-surface air temperature from multisource data},
  author={Du, Baoyu and Mao, Kebiao and Bateni, Sayed M and Meng, Fei and Wang, Xu-Ming and Guo, Zhonghua and Jun, Changhyun and Du, Guoming},
  journal={Remote sensing},
  volume={14},
  number={22},
  pages={5812},
  year={2022},
  publisher={MDPI}
}

@article{wang2025novel,
  title={A novel spatial prediction method integrating Exploratory Spatial Data Analysis into Random Forest for large scale daily air temperature mapping},
  author={Wang, Yuxue and Yin, Yue and Gao, Bingbo and Zeng, Yelu and Zhao, Yuanyuan and Chen, Ziyue and Feng, Quanlong and Xu, Hao and Yang, Jianyu},
  journal={IEEE Transactions on Geoscience and Remote Sensing},
  year={2025},
  publisher={IEEE}
}

@article{xu2018mapping,
  title={Mapping monthly air temperature in the Tibetan Plateau from MODIS data based on machine learning methods},
  author={Xu, Yongming and Knudby, Anders and Shen, Yan and Liu, Yonghong},
  journal={IEEE journal of selected topics in applied earth observations and remote sensing},
  volume={11},
  number={2},
  pages={345--354},
  year={2018},
  publisher={IEEE}
}

@article{kloog2015using,
  title={Using satellite-based spatiotemporal resolved air temperature exposure to study the association between ambient air temperature and birth outcomes in Massachusetts},
  author={Kloog, Itai and Melly, Steven J and Coull, Brent A and Nordio, Francesco and Schwartz, Joel D},
  journal={Environmental health perspectives},
  volume={123},
  number={10},
  pages={1053--1058},
  year={2015},
  publisher={NLM-Export}
}

@article{schwingshackl2024high,
  title={High-resolution projections of ambient heat for major European cities using different heat metrics},
  author={Schwingshackl, Clemens and Daloz, Anne Sophie and Iles, Carley and Aunan, Kristin and Sillmann, Jana},
  journal={Natural Hazards and Earth System Sciences},
  volume={24},
  number={1},
  pages={331--354},
  year={2024},
  publisher={Copernicus GmbH}
}

@article{tarek2021uncertainty,
  title={Uncertainty of gridded precipitation and temperature reference datasets in climate change impact studies},
  author={Tarek, Mostafa and Brissette, Fran{\c{c}}ois and Arsenault, Richard},
  journal={Hydrology and Earth System Sciences},
  volume={25},
  number={6},
  pages={3331--3350},
  year={2021},
  publisher={Copernicus Publications G{\"o}ttingen, Germany}
}

@article{sun2005air,
  title={Air temperature retrieval from remote sensing data based on thermodynamics},
  author={Sun, Y-J and Wang, J-F and Zhang, R-H and Gillies, RR and Xue, YYCB and Bo, Y-C},
  journal={Theoretical and applied climatology},
  volume={80},
  number={1},
  pages={37--48},
  year={2005},
  publisher={Springer}
}

@article{pape2004modelling,
  title={Modelling spatio-temporal near-surface temperature variation in high mountain landscapes},
  author={Pape, Roland and L{\"o}ffler, J{\"o}rg},
  journal={Ecological Modelling},
  volume={178},
  number={3-4},
  pages={483--501},
  year={2004},
  publisher={Elsevier}
}

@article{shen2020deep,
  title={Deep learning-based air temperature mapping by fusing remote sensing, station, simulation and socioeconomic data},
  author={Shen, Huanfeng and Jiang, Yun and Li, Tongwen and Cheng, Qing and Zeng, Chao and Zhang, Liangpei},
  journal={Remote Sensing of Environment},
  volume={240},
  pages={111692},
  year={2020},
  publisher={Elsevier}
}

@article{dai2024urban,
  title={Urban Air Temperature Prediction using Conditional Diffusion Models},
  author={Dai, Siyang and Liu, Jun and Cheung, Ngai-Man},
  journal={arXiv preprint arXiv:2412.13504},
  year={2024}
}

@article{lee2025estimating,
  title={Estimating Near-Surface Air Temperature from Satellite-Derived Land Surface Temperature Using Temporal Deep Learning: A Comparative Analysis},
  author={Lee, Jangho},
  journal={IEEE Access},
  year={2025},
  publisher={IEEE}
}

@article{cuomo2022scientific,
  title={Scientific machine learning through physics--informed neural networks: Where we are and what’s next},
  author={Cuomo, Salvatore and Di Cola, Vincenzo Schiano and Giampaolo, Fabio and Rozza, Gianluigi and Raissi, Maziar and Piccialli, Francesco},
  journal={Journal of Scientific Computing},
  volume={92},
  number={3},
  pages={88},
  year={2022},
  publisher={Springer}
}

@inproceedings{zhang2021physics,
  title={Physics-based iterative projection complex neural network for phase retrieval in lensless microscopy imaging},
  author={Zhang, Feilong and Liu, Xianming and Guo, Cheng and Lin, Shiyi and Jiang, Junjun and Ji, Xiangyang},
  booktitle={Proceedings of the IEEE/CVF Conference on Computer Vision and Pattern Recognition},
  pages={10523--10531},
  year={2021}
}

@inproceedings{gong2024physics,
  title={A physics-informed low-rank deep neural network for blind and universal lens aberration correction},
  author={Gong, Jin and Yang, Runzhao and Zhang, Weihang and Suo, Jinli and Dai, Qionghai},
  booktitle={Proceedings of the IEEE/CVF Conference on Computer Vision and Pattern Recognition},
  pages={24861--24870},
  year={2024}
}

@inproceedings{ji2025pomp,
  title={POMP: Physics-consistent Motion Generative Model through Phase Manifolds},
  author={Ji, Bin and Pan, Ye and Liu, Zhimeng and Tan, Shuai and Jin, Xiaogang and Yang, Xiaokang},
  booktitle={Proceedings of the Computer Vision and Pattern Recognition Conference},
  pages={22690--22701},
  year={2025}
}

@article{liu2022physics,
  title={Physics-informed hyperspectral remote sensing image synthesis with deep conditional generative adversarial networks},
  author={Liu, Liqin and Li, Wenyuan and Shi, Zhenwei and Zou, Zhengxia},
  journal={IEEE Transactions on Geoscience and Remote Sensing},
  volume={60},
  pages={1--15},
  year={2022},
  publisher={IEEE}
}

@article{shi2024physics,
  title={A physics-guided attention-based neural network for sea surface temperature prediction},
  author={Shi, Benyun and Feng, Liu and He, Hailun and Hao, Yingjian and Peng, Yue and Liu, Miao and Liu, Yang and Liu, Jiming},
  journal={IEEE Transactions on Geoscience and Remote Sensing},
  year={2024},
  publisher={IEEE}
}

@article{costa2025dani,
  title={DANI-NET: A Physics-Aware Deep Learning Framework for Change Detection Using Repeat-Pass InSAR},
  author={Costa, Giovanni and Guarnieri, Andrea Virgilio Monti and Parizzi, Alessandro and Rizzoli, Paola},
  journal={IEEE Transactions on Geoscience and Remote Sensing},
  year={2025},
  publisher={IEEE}
}

@inproceedings{liu2022deep,
  title={Deep decomposition for stochastic normal-abnormal transport},
  author={Liu, Peirong and Lee, Yueh and Aylward, Stephen and Niethammer, Marc},
  booktitle={Proceedings of the IEEE/CVF Conference on Computer Vision and Pattern Recognition},
  pages={18791--18801},
  year={2022}
}

@inproceedings{gruszczynski2025beyond,
  title={Beyond Blur: A Fluid Perspective on Generative Diffusion Models},
  author={Gruszczynski, Grzegorz and Meixner, Jakub and Wlodarczyk, Michal and Musialski, Przemyslaw},
  booktitle={Proceedings of the IEEE/CVF International Conference on Computer Vision},
  pages={17818--17827},
  year={2025}
}

@article{chai2020deep,
  title={Deep learning for irregularly and regularly missing 3-D data reconstruction},
  author={Chai, Xintao and Tang, Genyang and Wang, Shangxu and Lin, Kai and Peng, Ronghua},
  journal={IEEE Transactions on Geoscience and Remote Sensing},
  volume={59},
  number={7},
  pages={6244--6265},
  year={2020},
  publisher={IEEE}
}

@article{cuypers2021deep,
  title={Deep learning on construction sites: a case study of sparse data learning techniques for rebar segmentation},
  author={Cuypers, Suzanna and Bassier, Maarten and Vergauwen, Maarten},
  journal={Sensors},
  volume={21},
  number={16},
  pages={5428},
  year={2021},
  publisher={MDPI}
}

@article{hu2022deep,
  title={Deep depth completion from extremely sparse data: A survey},
  author={Hu, Junjie and Bao, Chenyu and Ozay, Mete and Fan, Chenyou and Gao, Qing and Liu, Honghai and Lam, Tin Lun},
  journal={IEEE Transactions on Pattern Analysis and Machine Intelligence},
  volume={45},
  number={7},
  pages={8244--8264},
  year={2022},
  publisher={IEEE}
}

@article{ke2021universal,
  title={Universal weakly supervised segmentation by pixel-to-segment contrastive learning},
  author={Ke, Tsung-Wei and Hwang, Jyh-Jing and Yu, Stella X},
  journal={arXiv preprint arXiv:2105.00957},
  year={2021}
}

@inproceedings{tseng2021learning,
  title={Learning to predict crop type from heterogeneous sparse labels using meta-learning},
  author={Tseng, Gabriel and Kerner, Hannah and Nakalembe, Catherine and Becker-Reshef, Inbal},
  booktitle={Proceedings of the IEEE/CVF Conference on Computer Vision and Pattern Recognition},
  pages={1111--1120},
  year={2021}
}

@article{moraes2025weakly,
  title={A Weakly Supervised and Self-Supervised Learning Approach for Semantic Segmentation of Land Cover in Satellite Images with National Forest Inventory Data},
  author={Moraes, Daniel and Campagnolo, Manuel L and Caetano, M{\'a}rio},
  journal={Remote Sensing},
  volume={17},
  number={4},
  pages={711},
  year={2025},
  publisher={MDPI}
}

@article{wang2020weakly,
  title={Weakly supervised deep learning for segmentation of remote sensing imagery},
  author={Wang, Sherrie and Chen, William and Xie, Sang Michael and Azzari, George and Lobell, David B},
  journal={Remote Sensing},
  volume={12},
  number={2},
  pages={207},
  year={2020},
  publisher={MDPI}
}

@article{kirkwood2022bayesian,
  title={Bayesian deep learning for spatial interpolation in the presence of auxiliary information},
  author={Kirkwood, Charlie and Economou, Theo and Pugeault, Nicolas and Odbert, Henry},
  journal={Mathematical Geosciences},
  volume={54},
  number={3},
  pages={507--531},
  year={2022},
  publisher={Springer}
}

@article{cutolo2024cloinet,
  title={CLOINet: ocean state reconstructions through remote-sensing, in-situ sparse observations and deep learning},
  author={Cutolo, Eugenio and Pascual, Ananda and Ruiz, Simon and Zarokanellos, Nikolaos D and Fablet, Ronan},
  journal={Frontiers in Marine Science},
  volume={11},
  pages={1151868},
  year={2024},
  publisher={Frontiers Media SA}
}

@inproceedings{archambault2023multimodal,
  title={Multimodal Unsupervised Spatio-Temporal Interpolation of satellite ocean altimetry maps},
  author={Archambault, Th{\'e}o and Filoche, Arthur and Charantonis, Anastase Alexandre and B{\'e}r{\'e}ziat, Dominique},
  booktitle={VISAPP},
  year={2023}
}

@inproceedings{li2023rainfall,
  title={Rainfall spatial interpolation with graph neural networks},
  author={Li, Jia and Shen, Yanyan and Chen, Lei and Ng, Charles Wang Wai},
  booktitle={International conference on database systems for advanced applications},
  pages={175--191},
  year={2023},
  organization={Springer}
}

@article{lou2024non,
  title={A non-uniform grid graph convolutional network for sea surface temperature prediction},
  author={Lou, Ge and Zhang, Jiabao and Zhao, Xiaofeng and Zhou, Xuan and Li, Qian},
  journal={Remote Sensing},
  volume={16},
  number={17},
  pages={3216},
  year={2024},
  publisher={MDPI}
}

@article{tian2021early,
  title={Early labeled and small loss selection semi-supervised learning method for remote sensing image scene classification},
  author={Tian, Ye and Dong, Yuxin and Yin, Guisheng},
  journal={Remote Sensing},
  volume={13},
  number={20},
  pages={4039},
  year={2021},
  publisher={MDPI}
}

@article{luo2025physics,
  title={Physics-informed neural networks for PDE problems: A comprehensive review},
  author={Luo, Kuang and Zhao, Jingshang and Wang, Yingping and Li, Jiayao and Wen, Junjie and Liang, Jiong and Soekmadji, Henry and Liao, Shaolin},
  journal={Artificial Intelligence Review},
  volume={58},
  number={10},
  pages={1--43},
  year={2025},
  publisher={Springer}
}

@article{grossmann2024can,
  title={Can physics-informed neural networks beat the finite element method?},
  author={Grossmann, Tamara G and Komorowska, Urszula Julia and Latz, Jonas and Sch{\"o}nlieb, Carola-Bibiane},
  journal={IMA Journal of Applied Mathematics},
  volume={89},
  number={1},
  pages={143--174},
  year={2024},
  publisher={Oxford University Press}
}

@article{cai2021physics,
  title={Physics-informed neural networks (PINNs) for fluid mechanics: A review},
  author={Cai, Shengze and Mao, Zhiping and Wang, Zhicheng and Yin, Minglang and Karniadakis, George Em},
  journal={Acta Mechanica Sinica},
  volume={37},
  number={12},
  pages={1727--1738},
  year={2021},
  publisher={Springer}
}

@article{muller2023deep,
  title={Deep pre-trained FWI: where supervised learning meets the physics-informed neural networks},
  author={Muller, Ana PO and Costa, Jess{\'e} C and Bom, Clecio R and Klatt, Matheus and Faria, Elisangela L and de Albuquerque, Marcelo P and de Albuquerque, Marcio P},
  journal={Geophysical Journal International},
  volume={235},
  number={1},
  pages={119--134},
  year={2023},
  publisher={Oxford University Press}
}

@inproceedings{yan2023st,
  title={ST-PINN: A self-training physics-informed neural network for partial differential equations},
  author={Yan, Junjun and Chen, Xinhai and Wang, Zhichao and Zhoui, Enqiang and Liu, Jie},
  booktitle={2023 International Joint Conference on Neural Networks (IJCNN)},
  pages={1--8},
  year={2023},
  organization={IEEE}
}

@inproceedings{liu2018picanet,
  title={Picanet: Learning pixel-wise contextual attention for saliency detection},
  author={Liu, Nian and Han, Junwei and Yang, Ming-Hsuan},
  booktitle={Proceedings of the IEEE conference on computer vision and pattern recognition},
  pages={3089--3098},
  year={2018}
}

@inproceedings{huang2019ccnet,
  title={Ccnet: Criss-cross attention for semantic segmentation},
  author={Huang, Zilong and Wang, Xinggang and Huang, Lichao and Huang, Chang and Wei, Yunchao and Liu, Wenyu},
  booktitle={Proceedings of the IEEE/CVF international conference on computer vision},
  pages={603--612},
  year={2019}
}

@article{li2018pyramid,
  title={Pyramid attention network for semantic segmentation},
  author={Li, Hanchao and Xiong, Pengfei and An, Jie and Wang, Lingxue},
  journal={arXiv preprint arXiv:1805.10180},
  year={2018}
}

@inproceedings{hassani2023neighborhood,
  title={Neighborhood attention transformer},
  author={Hassani, Ali and Walton, Steven and Li, Jiachen and Li, Shen and Shi, Humphrey},
  booktitle={Proceedings of the IEEE/CVF conference on computer vision and pattern recognition},
  pages={6185--6194},
  year={2023}
}

@inproceedings{long2015fully,
  title={Fully convolutional networks for semantic segmentation},
  author={Long, Jonathan and Shelhamer, Evan and Darrell, Trevor},
  booktitle={Proceedings of the IEEE conference on computer vision and pattern recognition},
  pages={3431--3440},
  year={2015}
}

@article{velickovic2017graph,
  title={Graph attention networks},
  author={Velickovic, Petar and Cucurull, Guillem and Casanova, Arantxa and Romero, Adriana and Lio, Pietro and Bengio, Yoshua and others},
  journal={stat},
  volume={1050},
  number={20},
  pages={10--48550},
  year={2017}
}

@inproceedings{xu2019spatial,
  title={Spatial-aware graph relation network for large-scale object detection},
  author={Xu, Hang and Jiang, Chenhan and Liang, Xiaodan and Li, Zhenguo},
  booktitle={Proceedings of the IEEE/CVF Conference on Computer Vision and Pattern Recognition},
  pages={9298--9307},
  year={2019}
}

@article{kavran2023graph,
  title={Graph neural network-based method of spatiotemporal land cover mapping using satellite imagery},
  author={Kavran, Domen and Mongus, Domen and {\v{Z}}alik, Borut and Luka{\v{c}}, Niko},
  journal={Sensors},
  volume={23},
  number={14},
  pages={6648},
  year={2023},
  publisher={MDPI}
}

@incollection{foken2024microclimatology,
  title={Microclimatology},
  author={Foken, Thomas and Mauder, Matthias},
  booktitle={Micrometeorology},
  pages={331--351},
  year={2024},
  publisher={Springer}
}

@article{su2002surface,
  title={The Surface Energy Balance System (SEBS) for estimation of turbulent heat fluxes},
  author={Su, Zhongbo},
  journal={Hydrology and earth system sciences},
  volume={6},
  number={1},
  pages={85--100},
  year={2002},
  publisher={Copernicus Publications G{\"o}ttingen, Germany}
}

@article{lagouarde1992daily,
  title={Daily sensible heat flux estimation from a single measurement of surface temperature and maximum air temperature},
  author={Lagouarde, J-P and McAneney, KJ},
  journal={Boundary-Layer Meteorology},
  volume={59},
  number={4},
  pages={341--362},
  year={1992},
  publisher={Springer}
}

@article{eliasof2023adr,
  title={Adr-gnn: advection-diffusion-reaction graph neural networks},
  author={Eliasof, Moshe and Haber, Eldad and Treister, Eran},
  journal={arXiv preprint arXiv:2307.16092},
  volume={108},
  year={2023}
}

@article{union2013advection,
  title={Advection diffusion equation models in near-surface geophysical and environmental sciences},
  author={Union, J Ind Geophys},
  journal={J. Ind. Geophys. Union},
  volume={17},
  pages={117--127},
  year={2013}
}

@incollection{clairambault2013reaction,
  title={Reaction-diffusion-advection equation},
  author={Clairambault, Jean},
  booktitle={Encyclopedia of Systems Biology},
  pages={1817--1817},
  year={2013},
  publisher={Springer}
}

@inproceedings{he2016deep,
  title={Deep residual learning for image recognition},
  author={He, Kaiming and Zhang, Xiangyu and Ren, Shaoqing and Sun, Jian},
  booktitle={Proceedings of the IEEE conference on computer vision and pattern recognition},
  pages={770--778},
  year={2016}
}

@article{verma2024climode,
  title={Climode: Climate and weather forecasting with physics-informed neural odes},
  author={Verma, Yogesh and Heinonen, Markus and Garg, Vikas},
  journal={arXiv preprint arXiv:2404.10024},
  year={2024}
}

@inproceedings{huang2023revisiting,
  title={Revisiting residual networks for adversarial robustness},
  author={Huang, Shihua and Lu, Zhichao and Deb, Kalyanmoy and Boddeti, Vishnu Naresh},
  booktitle={Proceedings of the IEEE/CVF conference on computer vision and pattern recognition},
  pages={8202--8211},
  year={2023}
}

@article{raissi2019physics,
  title={Physics-informed neural networks: A deep learning framework for solving forward and inverse problems involving nonlinear partial differential equations},
  author={Raissi, Maziar and Perdikaris, Paris and Karniadakis, George E},
  journal={Journal of Computational physics},
  volume={378},
  pages={686--707},
  year={2019},
  publisher={Elsevier}
}

@article{yaman2020self,
  title={Self-supervised learning of physics-guided reconstruction neural networks without fully sampled reference data},
  author={Yaman, Burhaneddin and Hosseini, Seyed Amir Hossein and Moeller, Steen and Ellermann, Jutta and U{\u{g}}urbil, K{\^a}mil and Ak{\c{c}}akaya, Mehmet},
  journal={Magnetic resonance in medicine},
  volume={84},
  number={6},
  pages={3172--3191},
  year={2020},
  publisher={Wiley Online Library}
}

@article{limoncella2025machine,
  title={A Machine Learning Model Integrating Remote Sensing, Ground Station, and Geospatial Data to Predict Fine-Resolution Daily Air Temperature for Tuscany, Italy},
  author={Limoncella, Giorgio and Feurer, Denise and Roye, Dominic and De Hoogh, Kees and De La Cruz, Arturo and Gasparrini, Antonio and Schneider, Rochelle and Pirotti, Francesco and Catelan, Dolores and Stafoggia, Massimo and others},
  journal={Remote sensing},
  volume={17},
  number={17},
  pages={3052},
  year={2025},
  publisher={MDPI}
}

@article{vedri2025empirical,
  title={Empirical methods to determine surface air temperature from satellite-retrieved data},
  author={Vedr{\'\i}, Joan and Nicl{\`o}s, Raquel and P{\'e}rez-Planells, Llu{\'\i}s and Valor, Enric and Luna, Yolanda and Estrela, Mar{\'\i}a Jos{\'e}},
  journal={International Journal of Applied Earth Observation and Geoinformation},
  volume={136},
  pages={104380},
  year={2025},
  publisher={Elsevier}
}

@article{zhang2024remotely,
  title={Remotely Sensed Estimation of Daily Near-Surface Air Temperature: A Comparison of Metop and MODIS},
  author={Zhang, Zhenwei and Li, Peisong and Zheng, Xiaodi and Zhang, Hongwei},
  journal={Remote Sensing},
  volume={16},
  number={20},
  pages={3754},
  year={2024},
  publisher={MDPI}
}

@article{noi2017comparison,
  title={Comparison of multiple linear regression, cubist regression, and random forest algorithms to estimate daily air surface temperature from dynamic combinations of MODIS LST data},
  author={Noi, Phan Thanh and Degener, Jan and Kappas, Martin},
  journal={Remote sensing},
  volume={9},
  number={5},
  pages={398},
  year={2017},
  publisher={MDPI}
}

@article{karagiannidis2025real,
  title={Real-Time Estimation of Near-Surface Air Temperature over Greece Using Machine Learning Methods and LSA SAF Satellite Products},
  author={Karagiannidis, Athanasios and Kyros, George and Lagouvardos, Konstantinos and Kotroni, Vassiliki},
  journal={Remote Sensing},
  volume={17},
  number={7},
  pages={1112},
  year={2025},
  publisher={MDPI}
}

@article{yang2017evaluation,
  title={Evaluation of MODIS land surface temperature data to estimate near-surface air temperature in Northeast China},
  author={Yang, Yuan Z and Cai, Wen H and Yang, Jian},
  journal={Remote Sensing},
  volume={9},
  number={5},
  pages={410},
  year={2017},
  publisher={MDPI}
}

@article{liu2017evaluating,
  title={Evaluating four remote sensing methods for estimating surface air temperature on a regional scale},
  author={Liu, Suhua and Su, Hongbo and Tian, Jing and Zhang, Renhua and Wang, Weizhen and Wu, Yueru},
  journal={Journal of Applied Meteorology and Climatology},
  volume={56},
  number={3},
  pages={803--814},
  year={2017}
}

@article{wang2017comparison,
  title={Comparison of spatial interpolation and regression analysis models for an estimation of monthly near surface air temperature in China},
  author={Wang, Mengmeng and He, Guojin and Zhang, Zhaoming and Wang, Guizhou and Zhang, Zhengjia and Cao, Xiaojie and Wu, Zhijie and Liu, Xiuguo},
  journal={Remote Sensing},
  volume={9},
  number={12},
  pages={1278},
  year={2017},
  publisher={MDPI}
}

@article{bouaziz2025wgast,
  title={WGAST: Weakly-supervised generative network for daily 10 m Land Surface Temperature estimation via spatio-temporal fusion},
  author={Bouaziz, Sofiane and Hafiane, Adel and Canals, Raphael and Nedjai, Rachid},
  journal={arXiv preprint arXiv:2508.06485},
  year={2025}
}
}


\end{document}